# Verified, not generated: expert-verified AI study materials and the distribution of learning gains in a university course

*Canh Thien Dang, King's Business School, King's College London*

*An Nguyen, King's Business School, King's College London*

**Highlights**

- AI-generated study materials were checked by a named tutor before release.
- Marks below the 60% classification boundary fell by 24.7 percentage points.
- About three-quarters of the 2.34-mark average gain arose in the bottom quintile.
- Dropping the lowest pre-period marks leaves the threshold effect; the mean weakens.
- Students read the verification label as a reason to engage, not to stop checking.

**Abstract**

Experimental studies of generative AI in education mostly report average effects, yet field evidence shows that AI can narrow attainment gaps or widen them. We argue that the direction depends on the judgement burden, the expertise a learner must supply to screen AI output before learning from it, and that expert verification before release moves this burden from students to an accountable tutor. We test the argument in a two-cohort difference-in-differences design in which one half of a compulsory first-year university economics course received AI-generated podcasts, FAQs and quiz-based study guides, produced with a source-grounded model and checked by a named graduate teaching assistant (170 students; 340 examination marks). Access was associated with a 2.34-mark advantage on a 50-mark component. The share of marks below the upper-second classification boundary fell by 24.7 percentage points relative to the counterfactual, effects were significant at every threshold from 23 to 31 marks and at none above, and roughly three-quarters of the average originated in the bottom quintile. The threshold estimate is robust to removing the lowest-scoring students from the pre-intervention cohort; the average effect is not. Interviews and feedback from 36 students indicate that the verification label gave students a reason to engage with AI-generated material without ending their scrutiny of it. Evaluations of AI learning resources that report only mean effects cannot detect whether the students the resources are meant to help are the ones who gain.

## 1. Introduction

Experimental research on generative AI in education has concentrated on average effects. A meta-analysis of experimental studies found that ChatGPT use improved academic performance on average (Deng et al., 2025), and comparisons of AI-generated with human-made teaching videos find that students learn about as much from either (Netland et al., 2025). Average effects, however, leave open the question that matters most for equity: which students gain. Field experiments outside education give opposite answers. In customer support, sales, professional writing and consulting, AI assistance raised the performance of weaker workers most and narrowed gaps (Brynjolfsson et al., 2025; Dell'Acqua et al., 2026; Liao et al., 2024; Noy & Zhang, 2023). Among Kenyan entrepreneurs given an AI business mentor, low performers did worse and high performers better, with no average effect (Otis et al., 2026). In high-school mathematics, unrestricted access to GPT-4 raised practice performance and then harmed examination performance once access was withdrawn, whereas a guarded tutor largely avoided the harm (Bastani et al., 2025). Educational technology has a record of benefiting the already advantaged: enrolment and completion in massive open online courses were concentrated among students from more affluent neighbourhoods (Hansen & Reich, 2015). Whether generative AI repeats that record depends on how it is deployed, and the literature has not identified which feature of deployment sets the direction.

We argue that the deciding feature is the judgement burden: the expertise a learner must supply to separate accurate, usable AI output from inaccurate or misaligned output before learning from it. Otis et al. (2026) located the source of their widening gap here. Entrepreneurs asked similar questions and received similar advice; what differed was which advice they chose to act on. For a student, the burden is the work of checking whether an AI explanation matches the course, whether a definition is correct and whether a worked answer can be trusted. That work draws on the subject knowledge, time and confidence that weaker students have least of. Where students carry the burden themselves, AI study support should help stronger students most. Where a tutor carries it before release, the support should reach weaker students without the screening cost.

Universities increasingly produce AI-generated study material themselves rather than leaving students to prompt general-purpose models. Evaluations of such material, including a contemporaneous study of AI-generated podcasts in a US business course (Rezvani, 2026), report average outcomes. Institutional responses to generative AI, however, have concentrated on regulating what students do with it, through acceptable-use rules, disclosure requirements and detection (Moorhouse et al., 2023; Nguyen et al., 2025). Quality assurance of AI material that a course team issues to its own students has received less attention, and such material often carries no named person answerable for its accuracy.

We study a course in which the burden was moved explicitly. In 2024/25, one half of a compulsory first-year economics course at a UK university received AI-generated podcasts, FAQs and quiz-based study guides, produced with a source-grounded model and checked by a named graduate teaching assistant (GTA) before release under the label "AI-generated and academically verified". The other half received nothing comparable, and in the previous year neither half did. We address three research questions:

**RQ1.** Is access to expert-verified AI-generated study materials associated with higher examination performance?

**RQ2.** Where in the attainment distribution do any gains arise?

**RQ3.** How do students construe and use AI-generated materials that carry a named verifier?

Access was associated with a 2.34-mark advantage over the parallel-trends counterfactual on a 50-mark component. About three-quarters of that average originates in the bottom quintile of the mark distribution, and effects are significant at every threshold from 23 to 31 marks, a band ending at the upper-second classification boundary, and at none above it. The threshold result is robust to removing the lowest-scoring students from the pre-intervention cohort; the average effect is not. Students described the verification label as a reason to engage with material they would otherwise have doubted, without abandoning scrutiny.

We make three contributions. First, we introduce judgement burden as a learner-level construct that reconciles the levelling and widening results on generative AI and predicts where in the attainment distribution verified AI materials should help. Second, we provide quasi-experimental evidence, with a contemporaneous untreated comparison within the same course, that expert-verified AI study materials are associated with gains concentrated among the lowest-attaining students, in a deployment where students never prompted the model. Third, we show that the average effect of an AI learning resource can be statistically fragile while its effect at a consequential grade boundary is robust, and we propose distributional reporting standards for evaluations of AI in education.

## 2. Theoretical background

### *2.1. Generative AI and the distribution of learning gains*

The levelling results share a feature: either the AI carried expertise that weaker users lacked, or it worked on tasks where its output was reliably good. Brynjolfsson et al. (2025) studied an assistant trained on the conversations of successful support agents; it transmitted their practices to novices, whose productivity rose most. Inexperienced e-commerce sales agents gained nearly six times as much from AI augmentation as experienced agents (Liao et al., 2024). ChatGPT compressed the productivity distribution in professional writing by benefiting less able writers more (Noy & Zhang, 2023), and compressed it among consultants on tasks within the model's capability (Dell'Acqua et al., 2026).

The widening and harmful results share the opposite feature. On a task outside the model's capability, consultants using AI were less likely to reach correct answers than those working without it (Dell'Acqua et al., 2026). The Kenyan entrepreneurs received open-ended advice and had to decide which of it to act on (Otis et al., 2026). Students with unrestricted GPT-4 access could obtain answers without working through problems, and their learning suffered (Bastani et al., 2025). In each case the user, not the deploying organisation, decided what to accept from the AI and how far to rely on it.

*2.2. AI-generated podcasts and study materials*

Podcasts have been studied as revision resources in higher education for nearly two decades. Students judged podcast revision lectures more effective than their textbooks and more efficient than their own notes (Evans, 2008), although a review found little research evidence on podcasting's effect on attainment (Heilesen, 2010). Generative AI changed the cost of production. Tools such as Google NotebookLM turn a set of course documents into a conversational audio episode, a set of questions and a summary in minutes, and evaluations have begun to assess the quality of such podcasts as a means of communicating specialist content (Desmedt et al., 2025). AI-assisted audio learning has been associated with higher academic achievement through motivation and reading engagement (Jafarian & Kramer, 2025), and experimental comparisons of AI-generated with human-made or recorded instructional videos find learning outcomes of similar magnitude (Netland et al., 2025; Xu et al., 2025).

In a semester-long field setting, Rezvani (2026) compared a semester of a US Management Information Systems course in which NotebookLM podcasts and study guides were available with one in which they were not, and found higher exam scores in the podcast semester and a positive association between podcast engagement and course outcomes.

Two features of this literature frame our study. First, quality control is present but outside the design. In the video experiments the researchers fixed what the materials said; in classroom deployments instructors review outputs before release. In routine use, the accuracy of AI-generated material varies, and someone must check it, but who checks it, and whether students know, has not been treated as a variable. Second, outcomes are reported as averages, so the evidence cannot yet say which students the materials help.

*2.3. Judgement burden*

We define judgement burden as the expertise and effort a learner must supply to distinguish usable from unusable AI output before learning from it. It has two components: the variability of output quality, which determines how much checking is needed, and the location of responsibility for checking, which determines who must do it.

Research on credibility describes what the burden involves. Users evaluating online information weigh accuracy, authority, objectivity, currency and coverage, and most rely on heuristics rather than systematic checking because systematic checking is costly (Metzger, 2007). College students assessing AI-generated information for academic tasks perform this screening themselves, comparing outputs with course materials and their own knowledge, and report uncertainty about when AI output can be trusted (Choi et al., 2025). Research on how students integrate ChatGPT content into their academic writing links whether they check it to their epistemic beliefs and metacognitive accuracy (Urban et al., 2025), which are themselves unevenly distributed across students. Research on reliance shows two failure modes. Users of decision aids over-rely on advice and miss its errors, and this complacency grows with the aid's apparent reliability (Parasuraman & Manzey, 2010). People also abandon an algorithmic adviser faster than a human one after seeing the same error (Dietvorst et al., 2015), although they often prefer algorithmic advice before seeing any error (Logg et al., 2019). Trust in AI depends on transparency and reliability among other antecedents (Glikson & Woolley, 2020).

Judgement burden is related to, and distinct from, cognitive load. Cognitive load theory distinguishes the intrinsic load of the material from extraneous load imposed by how it is presented (Sweller et al., 2019). Checking an AI explanation for accuracy is extraneous to learning the content: it consumes working memory and time without advancing understanding of the concept being checked. Judgement burden names the source of that extraneous load in AI-generated material and adds the question of who bears it. Identical content imposes a heavy burden when released unchecked and a light one when released verified.

The burden is regressive. Checking draws on prior knowledge of the subject, which weaker students lack by definition, and on time and confidence, which students with paid work, long commutes or study in a second language have less of. The outcomes students draw from digital resources track the resources they bring to them (van Deursen & Helsper, 2015), and even structured support converts into benefits unevenly (Kim et al., 2026). When AI material varies in quality and students must check it, stronger students should extract its benefits while weaker students either act on its errors or discard it. When a tutor checks it before release, the students who lacked the resources to check it gain most, because the material supplies what they lack, while for stronger students it largely duplicates what they already have. Fig. 1 locates the studies reviewed above on these two dimensions.

The account implies two channels. Supplementary material in accessible formats can help weaker students whether or not AI produced it; we call this the access channel. AI-generated material adds a second channel that runs the other way: released unchecked, it imposes a burden that weaker students are least able to carry, diverting its benefits toward stronger students. Verification does not create the access channel. It closes the diversion channel, so that the access channel can operate for the students it is meant to reach.

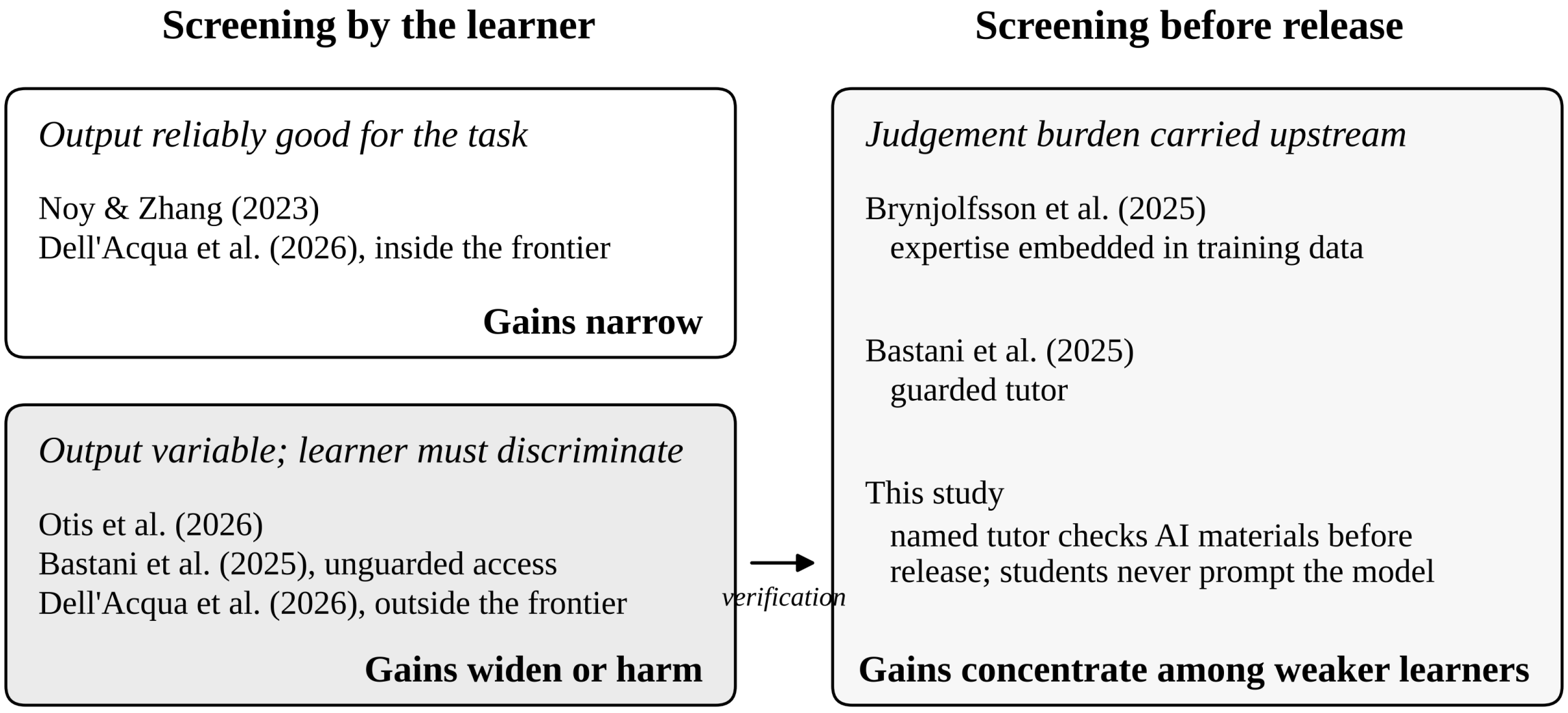

**Fig. 1.** Judgement burden and the distribution of gains from generative AI. Where users must screen variable output, gains widen; where screening is performed upstream, gains concentrate among weaker users.

### *2.4. Expert-verified AI study materials*

We define expert-verified AI study materials as resources with three jointly necessary features: they are generated by a source-grounded model from the course's own documents; they are checked by a named expert before release; and they are labelled to disclose both their AI origin and their verification. Source-grounding bounds the verifier's task. Retrieval-augmented generation conditions output on a defined document set (Lewis et al., 2020), so the verifier checks fidelity to known sources rather than an open-ended space of claims. It reduces but does not eliminate hallucination (Ji et al., 2023), which is why named verification is required. Disclosure makes both the machine origin and the human responsibility visible to students.

Verification does three jobs in this arrangement. It *screens*, removing errors and misalignment before students see them, which is the mechanism that relocates the judgement burden. It *signals*: a named verifier gives students a low-cost reason to engage while placing the grounds of trust in human checking rather than in the fluency of the text. Credibility cues supplied by an interface need not replace users' own evaluation; credibility indicators for health webpages helped users assess credibility and prompted them to validate their initial judgements (Zhang et al., 2023). A verification label can work the same way, as a prompt to scrutiny rather than a substitute for it. And it is *labour*: verification relocates work rather than removing it, demands subject expertise, and depends for its signalling value on a relationship between verifier and students. Work of this kind tends to go unrecorded in academic workload models (Fleming & Harley, 2024).

### *2.5. Hypotheses*

The account yields two testable predictions and one interpretive expectation.

**H1.** Access to expert-verified AI study materials is associated with higher examination performance.

**H2.** Gains are concentrated in the lower tail of the attainment distribution, and attenuate toward the upper tail.

H2 carries the theoretical weight. The access channel alone could also produce lower-tail gains, so H2 does not distinguish verification from new material; it could, however, fail. Gains concentrated at the median or in the upper tail would disconfirm it even with an identical average. For RQ3 we expected that students would describe the verification label as a reason to engage that did not end their own checking.

## 3. Method

### *3.1. Setting and participants*

The study took place in a compulsory first-year Principles of Economics course at a UK university business school. The course has two five-week halves, microeconomics (Part 1) and macroeconomics (Part 2), each taught by the same lecturer in both observed years. A single two-hour terminal examination allocates 50 marks to each half and determines the whole module mark; there is no coursework component. The same GTA led tutorials in both halves and both years, and marking used the same rubric and marker. Each cohort comprised 85 students, and no student appears in both cohorts. In UK higher education, degree classes are defined by percentage bands: 40% is the pass mark, 50% the lower-second boundary, 60% the upper-second boundary on which most graduate employers and postgraduate programmes screen applicants, and 70% the first-class boundary. With 50 marks per half, these fall at component marks of 20, 25, 30 and 35. Studies of student performance in undergraduate economics have used course examination data in a similar way to examine how learning resources relate to attainment (Ipinnaiye & Risquez, 2024).

### *3.2. Intervention*

In 2024/25, the Part 2 lecturer uploaded lecture slides, readings and assessment guidance to Google NotebookLM and generated, for each weekly topic, a ten-minute conversational podcast, a set of FAQs, and a study guide with comprehension questions and model answers. NotebookLM was chosen because it generates only from uploaded documents, produces several formats from one source set, and operated under institutional arrangements in which uploaded content is not used for model training. The GTA checked every output against the source materials: whether definitions matched the slides, whether models and diagrams used the course's terminology, whether examples suited the level, and whether quiz answers could be justified from the documents. Where outputs were vague or misaligned, the GTA revised the prompts or regenerated the sections rather than accepting the first draft. First drafts were usable but not releasable; the materials students received depended on the correction step.

Materials were released free through the virtual learning environment, labelled "AI-generated and academically verified", as a supplement to unchanged lectures, tutorials and readings. The multimodal formats followed Universal Design for Learning principles, offering several means of access for students whose circumstances limit conventional study (CAST, 2018). Use was voluntary and carried no marks. Part 1 received no comparable materials in 2024/25, and neither half did in 2023/24, so the intervention occupies one cell of a two-by-two design.

*3.3. Outcome measure and identification*

The outcome is each student's mark out of 50 on each half of the terminal examination. We estimate the following specification using OLS:

$$Mark_{ijt} = \alpha + \beta(Part2_j \times Post_t) + \gamma Post_t + \delta Part2_j + \varepsilon_{ijt} \quad (1)$$

where Mark is the score of student $i$ on half $j$ in year $t$, and β is the difference-in-differences estimate (Angrist & Pischke, 2009). Because every treated-cell student had access regardless of use, β is an intention-to-treat effect of access. We report heteroskedasticity-robust (HC1), student-clustered and within-student specifications.

The identifying assumption is that the Part 2 minus Part 1 gap would have evolved similarly across cohorts without the intervention. With one pre-intervention cohort this cannot be tested directly. The examination papers constrain the most direct threat to it. Because the design compares component marks across years, a change in the relative difficulty of the two halves between papers would by itself produce a nonzero estimate. The paper sat in 2024/25 was the resit paper drafted alongside the 2023/24 main paper, in the same round, by the same examiner, against the same specification, and moderated with it. No resit sitting took place, so the paper was unseen by the treated cohort. The relative difficulty of the two halves was therefore fixed at one point in time under one set of judgements, and the paper predates the intervention.

Two threats remain. The cohorts share no students, so cohort differences in ability cannot be netted out. The more serious threat is effort substitution. Both halves are examined together, so revision takes place after the Part 2 materials exist, and students with a fixed revision budget may have moved time toward the half with new resources. This would raise Part 2 marks and depress Part 1 marks without any gain in learning. Section 4.4 presents the evidence bearing on it.

*3.4. Distributional analysis*

Because H2 concerns heterogeneity, we supplement the mean regression with three distributional estimators. We compare variances with the Brown–Forsythe test. We estimate linear probability versions of the equation with an indicator for a mark below threshold $c$ as the outcome, for every $c$ from 20 to 45. We estimate quantile difference-in-differences at deciles, with 95% intervals from 1,000 bootstrap replications that resample students within cohort so that each student's two marks stay together. To assess whether results depend on a few extreme marks, we re-estimate after removing the lowest-scoring students in the pre-intervention treated cell and after winsorising marks from below.

*3.5. Qualitative data*

The design is sequential and explanatory (Venkatesh et al., 2013): the quantitative analysis establishes the distributional pattern, and the qualitative data are used to interpret mechanisms behind it. In the weeks after the examination, a research assistant with no teaching or marking role conducted brief, informal interviews and gathered feedback from 36 students, purposively selected for variation in prior performance, linguistic background, paid work and self-reported neurodiversity. The study received institutional ethics approval, participation was voluntary and unrelated to grading, and pseudonyms are used. The research assistant recorded contemporaneous notes capturing students' phrasing as closely as possible; these notes, not verbatim transcripts, are the source of the excerpts below, lightly edited for readability. We organised the notes inductively (Braun & Clarke, 2006) and then compared the themes with the hypotheses. The material illustrates mechanisms and supports no finding on its own.

## 4. Results

*4.1. Average effect (RQ1)*

Table 1 reports cell means and the main estimates. The coefficient on Part 2 × Post is 2.341 in all three specifications, identical by construction in a balanced two-by-two panel, with $p = 0.045$ (HC1), 0.013 (student-clustered) and 0.013 (within-student). The estimate is 0.33 standard deviations of the pre-intervention Part 2 distribution and 0.40 of the pooled pre-intervention distribution. Effect sizes on researcher- or course-designed assessments are typically larger than on standardised tests (Kraft, 2020), so these magnitudes should not be compared with benchmarks from standardised achievement measures. Decomposition qualifies the interpretation. Part 2 marks rose by 0.73, which is not significant ($p = 0.441$), while Part 1 marks fell by 1.61 ($p = 0.021$). The treated half held its level in a year when the untreated half declined, so the estimate describes protection from a decline rather than a rise. H1 receives qualified support, subject to the sensitivity results in Section 4.3.

**Table 1**
Access to verified AI study materials is associated with a 2.34-mark advantage over the parallel-trends counterfactual, reflecting a decline in the untreated half rather than a rise in the treated half.

*Panel A. Examination marks by cohort and half (out of 50)*

| Cohort | Half | N | Mean | SD |
|---|---|---|---|---|
| 2023/24 | Part 1 (untreated) | 85 | 35.87 | 4.32 |
| 2023/24 | Part 2 (untreated) | 85 | 34.12 | 7.09 |
| 2024/25 | Part 1 (untreated) | 85 | 34.26 | 4.67 |
| 2024/25 | Part 2 (treated) | 85 | 34.85 | 5.03 |

*Panel B. Difference-in-differences estimates*

| Variable | (1) HC1 | (2) Student-clustered | (3) Within-student |
|---|---|---|---|
| Part 2 × Post | 2.341 (1.169) | 2.341 (0.947) | 2.341 (0.946) |
| *p*-value | 0.045 | 0.013 | 0.013 |
| Post | −1.612 (0.690) | −1.612 (0.691) | absorbed |
| Part 2 | −1.753 (0.901) | −1.753 (0.780) | −1.753 (0.779) |
| Observations | 340 | 340 | 340 |

*Notes:* Standard errors in parentheses. Column 3 regresses each student's Part 2 minus Part 1 difference on Post.

### *4.2. Distributional effects (RQ2)*

A difference-in-differences coefficient identifies an average treatment effect and is uninformative about whether students at different levels of attainment benefited to different degrees. Table 2 and Fig. 2 report the tests of H2.

**Table 2**
Gains are confined to marks below the upper-second boundary and concentrated in the bottom quintile.

| Test | Estimate | Inference |
|---|---|---|
| *Dispersion* | | |
| Part 2 standard deviation | 7.09 → 5.03 | Brown–Forsythe $p = 0.004$ |
| Part 1 standard deviation | 4.32 → 4.67 | Brown–Forsythe $p = 0.873$ |
| *Threshold difference-in-differences* | | |
| Below 20 (pass mark, 40%) | −2.4 pp | $p = 0.155$; few observations below |
| Below 25 (lower second, 50%) | −12.9 pp | $p = 0.008$ |
| Below 30 (upper second, 60%) | −24.7 pp | $p = 0.001$ |
| Below 35 (first, 70%) | −2.4 pp | $p = 0.828$ |
| Every threshold 20 to 45 | Significant at every threshold 23 to 31; none above | Fig. 2a |
| *Quantiles and decomposition* | | |
| 10th percentile | +11.0 marks | 95% CI 5.0 to 13.0 |
| 20th percentile | +5.0 marks | 95% CI −1.0 to 10.0 |
| Bottom-quintile contribution | +8.9 marks | ≈ 75% of the average |
| *Alternative explanations* | | |
| Component correlation | 0.284 → 0.482 | Fisher $z = 1.50$, $p = 0.135$ |
| Kolmogorov–Smirnov, Part 2 | $D = 0.153$ | $p = 0.274$ |

*Notes:* pp = percentage points. Threshold estimates from linear probability models with HC1 errors. Quantile intervals from 1,000 student-level bootstrap resamples.

Dispersion changed in the treated half alone. The Part 2 standard deviation fell from 7.09 to 5.03 (Brown–Forsythe $p = 0.004$), while Part 1's moved from 4.32 to 4.67 ($p = 0.873$). The treated half's interdecile range narrowed from 19.6 to 11.6 marks; the untreated half's widened from 10.0 to 11.6.

Threshold estimates locate the change. At the upper-second boundary, the number of treated-half marks below 30 fell from 20 to 9 while the number of untreated-half marks below 30 rose from 6 to 16. In shares, the treated half moved from 23.5% to 10.6% and the untreated half from 7.1% to 18.8%, a difference-in-differences of −24.7 percentage points ($p = 0.001$), equivalent to 21 students in a cohort of 85. At the lower-second boundary the estimate is −12.9 points ($p = 0.008$); at the first-class boundary it is −2.4 points ($p = 0.828$). Re-estimating at every threshold from 20 to 45 marks (Fig. 2a) shows estimates significant at every threshold from 23 to 31, largest in absolute value at 29 (−25.9 points), and significant at no threshold above 31. The effect occupies a bounded band whose upper edge lies at the upper-second boundary.

The lower edge of the band is set by sparse data rather than by the mechanism. Only thirteen marks in the dataset fall below 23, seven below 22 and four below 21, most of them in a single cell, so estimates there are unstable: significant at 21, not at 22 ($p = 0.251$). At the pass mark the estimate is −2.4 points and not significant ($p = 0.155$). No student failed the course in either cohort; module means were 70.0 and 69.1 out of 100, and the lowest totals were 45 and 47. The estimated effect therefore concerns degree classification, and this cohort carries no information about failure.

Quantile estimates are shown in Fig. 2b. The gain is 11.0 marks at the 10th percentile (95% interval 5.0 to 13.0) and 5.0 marks at the 20th (−1.0 to 10.0), and estimates from the 30th percentile upward are close to zero. Only the 10th-percentile estimate is bounded away from zero. Decomposing the average by quintile of each cohort's distribution attributes roughly three-quarters of the 2.34 marks to the bottom quintile, whose difference-in-differences is 8.9 marks, and almost none to the top three. An omnibus Kolmogorov–Smirnov test comparing the treated half's two cohort distributions is not significant ($D = 0.153$, $p = 0.274$). That result is expected when change is confined to one tail, because the statistic is dominated by the centre of the distribution, where nothing moved. With 85 students per cell, tail estimates rest on small counts. The dispersion, threshold and quantile results agree in direction and location, and that agreement is the basis for our interpretation. H2 is supported.

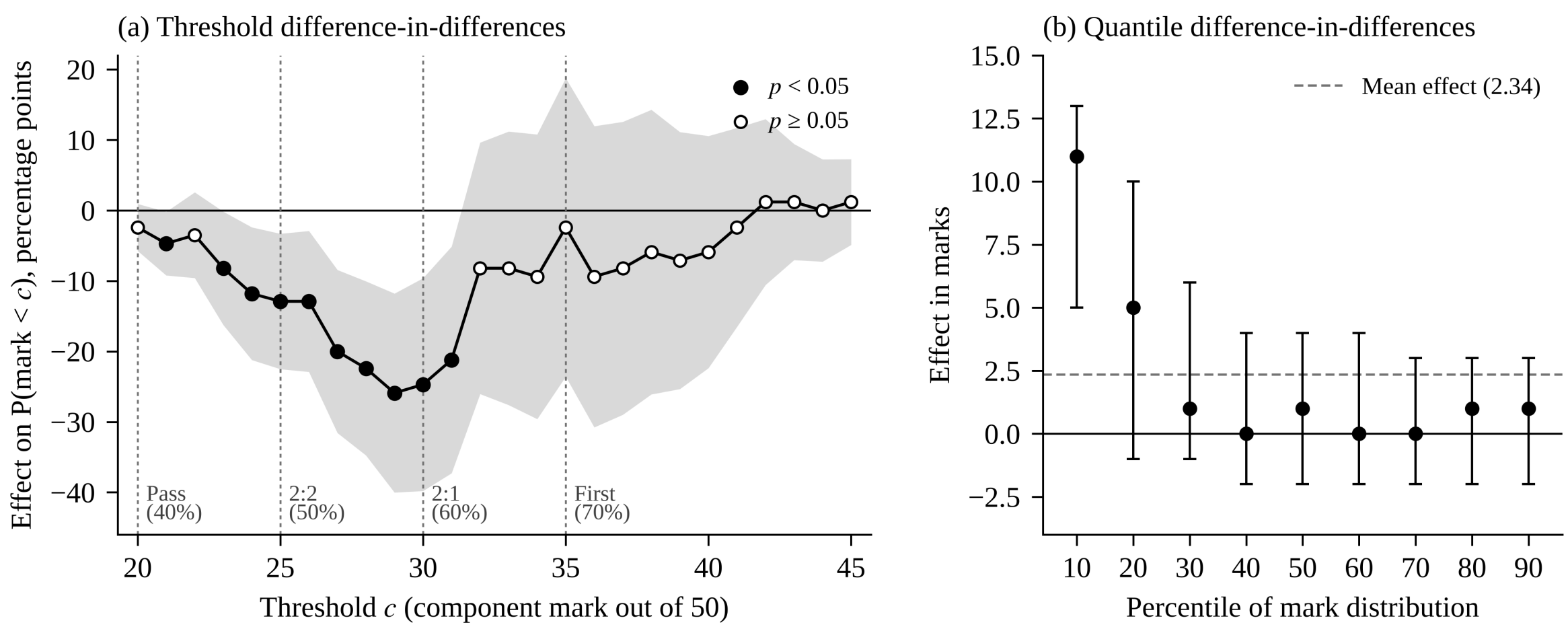

**Fig. 2.** Distributional estimates. (a) Difference-in-differences on the probability of scoring below each threshold, with 95% intervals; UK classification boundaries marked. (b) Quantile difference-in-differences with 95% intervals from 1,000 student-level bootstrap resamples.

### *4.3. Sensitivity to the lowest pre-intervention marks*

A lower-tail effect estimated from 85 students per cell raises a concern: a handful of very low marks in the 2023/24 Part 2 cell could generate both the average effect and the threshold results. Table 3 reports estimates after removing the lowest-scoring students in that cell, together with both of their marks, and after winsorising all marks from below.

**Table 3**
The reduction in marks below the upper-second boundary is robust to removing the lowest pre-intervention marks; the average effect attenuates.

| Sample | Observations | Average effect (marks) | *p* | Below 30 (pp) | *p* |
|---|---|---|---|---|---|
| Full sample | 340 | 2.34 (1.17) | 0.045 | −24.7 (7.7) | 0.001 |
| Drop 2 students below 20 in 2023/24 Part 2 | 336 | 1.94 (1.15) | 0.093 | −22.7 (7.6) | 0.003 |
| Drop lowest 3 in 2023/24 Part 2 | 334 | 1.88 (1.14) | 0.099 | −22.9 (7.5) | 0.002 |
| Drop lowest 5 in 2023/24 Part 2 | 330 | 1.56 (1.13) | 0.166 | −20.7 (7.5) | 0.006 |
| Drop lowest 8 in 2023/24 Part 2 | 324 | 1.41 (1.09) | 0.198 | −19.9 (7.2) | 0.006 |
| Winsorised at pooled 5th percentile (23) | 340 | 2.14 (1.13) | 0.057 | −24.7 (7.7) | 0.001 |
| Winsorised at pooled 10th percentile (28) | 340 | 1.34 (0.98) | 0.173 | −24.7 (7.7) | 0.001 |

*Notes:* HC1 standard errors in parentheses. With student-clustered errors, the average-effect *p*-values for the four drop rows are 0.035, 0.041, 0.085 and 0.127. Dropped students are removed with both of their marks. Winsorising below 30 leaves the below-30 indicator unchanged. Removing low marks from the pre-intervention treated cell only biases estimates toward zero.

The two estimands respond differently. The average effect attenuates as low marks are removed. Dropping the three lowest-scoring students reduces it from 2.34 to 1.88 marks (HC1 $p = 0.099$; student-clustered $p = 0.041$), dropping five reduces it to 1.56 (clustered $p = 0.085$), and dropping eight reduces it to 1.41 (clustered $p = 0.127$). Winsorising at the pooled 5th percentile (23 marks) leaves 2.14 ($p = 0.057$); winsorising at the 10th (28 marks) leaves 1.34 ($p = 0.173$). The threshold estimate at the upper-second boundary is stable when students are removed, lying between −19.9 and −22.9 percentage points with $p \leq 0.006$ in every case; winsorising below 30 leaves it unchanged by construction.

The asymmetry follows from what each estimand measures. The average weights each mark by its distance from the mean, so a few marks near 20 carry large weight; removing them, or pulling them up by winsorising, removes much of the variation that produces the average effect. The threshold estimate counts students on either side of 30 marks, so it does not depend on how far below 30 the lowest marks lie. The exercise is stringent by construction, because it removes low marks from the pre-intervention treated cell only, which biases the estimates toward zero. The statistical significance of the average effect therefore rests on a small number of students, whereas the reduction in the share of marks below the upper-second boundary does not. H1 should be read in that light; H2 is a claim about the second quantity.

*4.4. Alternative explanations*

Effort substitution predicts a specific signature. If students moved revision time from Part 1 to Part 2, the within-student correlation between the two component marks should weaken in the treated cohort, because gains in one half would come at the expense of the other. The correlation instead rose from 0.284 to 0.482, although the change is not significant (Fisher $z = 1.50$, $p = 0.135$). Substitution would also be available to students across the distribution, whereas the gains are confined to the lower tail. Neither observation excludes substitution, and it remains an alternative to the treatment interpretation. The distributional pattern bears similarly on novelty effects: a motivational response to new resources should raise marks broadly, not within a band that ends at 31 marks.

A different alternative concerns mechanism rather than identification. The treated half received both verification and new multimodal material, and additional material in accessible formats could by itself help weaker students through the access channel. The quiz-based study guides add a specific version of this channel. Attempting a question before reading its answer is retrieval practice, which improves retention more than restudying (Roediger & Karpicke, 2006), and feedback after an attempt strengthens the effect and corrects errors (Butler & Roediger, 2008). Because our guides included model answers, students could use them either for self-testing with feedback or for rereading, and we cannot observe which. The lower-tail pattern is consistent with these accounts as well as with ours, and the design cannot apportion the effect between them. The judgement-burden account adds a prediction about the counterfactual the design lacks: the same material released unverified should produce smaller lower-tail gains, or none, because weaker students would bear the checking. The field evidence on unguarded use points that way (Bastani et al., 2025; Otis et al., 2026), but our data do not test it.

*4.5. Students' accounts of verified materials (RQ3)*

Students' accounts followed a consistent sequence: scepticism, then reassurance from the label, then patterns of use. Initial caution was reasoned rather than technophobic: "I've used ChatGPT for quick summaries before and it's got stuff wrong. So I was like, not sure I'd trust that for my exam" (Jasper); "My sister's a computer scientist and she's always going on about how AI makes up stuff. So I opened the FAQ document expecting to find something silly" (Priya). The label changed students' stance toward the material: "Up to then I was half-listening, waiting for a mistake. After that, I relaxed and actually learned" (Leila). The trust it conferred was social, attached to a person students knew: "I don't trust an AI. But I do trust my GTA. If she says it's okay, then I'm okay" (Matteo); "The moment I saw the GTA's name on the verification note, someone I'd actually spoken to in tutorials, I thought, oh, this is serious" (Grace). It did not end scrutiny: "The verification made me more critical, not less. Instead of asking, is this right, my questions became about depth" (Hamza). Hamza's account resembles the response to supplied credibility cues, which prompted users to validate their initial judgements (Zhang et al., 2023).

Several students described being relieved of the judgement burden in terms close to extraneous load. "I didn't have that nagging worry that some fact was made up. I could focus on understanding rather than fact-checking" (Ravi). "Knowing the content was verified meant I could listen without constantly stopping to wonder if that was really a term" (Declan, who reported dyslexia). The accounts that aligned most closely with the quantitative pattern came from students describing constraints on conventional study: a second language ("In lectures, I miss maybe 20% because people speak fast. The podcast I could pause, replay, slow down", Aisha), a long commute (Noah), reluctance to ask questions in tutorials (Daniel), and paid work ("Having everything on my phone, offline, with audio and text, levelled the playing field for me", Ruby). Constraints of this kind limit the time, language and confidence that checking consumes, which is the mechanism H2 proposes; the accounts cannot, however, link these students to particular positions in the mark distribution. Some students carried the verification norm beyond the course: “A friend sent me a ChatGPT summary and I asked, did you check that against the original?” (Freya). We have no evidence that this habit persisted.

*4.6. Take-up and the cost of verification*

Virtual learning environment records show that each podcast reached more than 70 unique listeners, about 82% of the cohort, and that FAQs and study guides were opened by about 65% and 60% of students. Because use is self-selected and individual usage logs were not retained, these figures establish take-up, not effects of use. Producing and verifying each weekly set took about 30 minutes, most of it spent checking rather than generating, and about 2.5 hours across five weeks: roughly 1.8 minutes of staff time per student. Verification time depends on the volume of material rather than the number of students, so the same 2.5 hours would serve a cohort of 300 at about half a minute per student. Against unverified generation, verification is the entire additional cost, since generation itself is almost free. A school-randomised trial found that ChatGPT cut teachers’ weekly lesson-planning time by 31% with no detectable loss of quality (Roy et al., 2024), without any verification step. These figures rest on one verifier’s estimates in one course, without an independent time audit.

## 5. Discussion

*5.1. Who gains from AI study materials*

Our first contribution is an account of when generative AI narrows attainment gaps. Judgement burden treats the levelling and widening results as outcomes of a single construct: where learners must check variable output, gains accrue to those with the knowledge to check it, and where checking is done before release, gains accrue to those who lacked that knowledge. Our results extend the levelling evidence to a deployment the field experiments did not examine. The students never prompted the model. They received content a tutor had already checked, and gains concentrated in the lower tail as the account predicts. The claim concerns verified AI materials as a package, since our design cannot separate verification from the material it accompanied.

The account also clarifies what favourable cases have in common. Brynjolfsson et al. (2025) placed judgement in the training data, by building the assistant on the conversations of successful agents. Bastani et al. (2025) placed it in the design of a tutor that withheld answers. Our setting placed it in a named tutor who checked content before release. Each form of upstream checking carries a cost that someone bears; the judgement burden is relocated rather than removed. Verification by a named tutor makes that cost visible, which allows it to be budgeted and also exposes it to being cut.

The location of the effect also bears on what the materials taught. The podcasts, FAQs and study guides were built from the course’s slides and readings, and the quiz questions tested definitions, standard models and their diagrams. Under the marking scheme, accurate statement and use of the standard models carries an answer into the upper-second range, while first-class marks require applying those models to unfamiliar cases and evaluating them, which material grounded in the existing slides does not supply. An effect band ending at the upper-second boundary is what one would expect if the materials consolidated core content for students who had not yet secured it and added little for students who had. This reading rests on the content of the materials and the marking scheme rather than on item-level evidence, which the examination data do not allow, but it implies that the reach of verified AI materials up the distribution depends on what they are built from. Materials generated from worked applications and past evaluative answers could, on the same account, move a different part of the distribution.

For research on AI-generated learning resources, the account implies a different comparison from the one most studies make. Comparing AI-generated with human-made material, with accuracy held fixed by the researchers, establishes that AI material can teach as well (Netland et al., 2025), and evaluations of AI-generated podcasts in which the instructor reviewed outputs establish that such material can accompany higher examination scores (Rezvani, 2026). It does not tell educators what happens when accuracy is not held fixed, which is the condition of routine use. The comparison our account calls for is between verified and unverified releases of the same AI material, with outcomes examined across the attainment distribution.

### *5.2. Comparison with other evaluations of AI-generated podcasts*

Our results agree in direction with a contemporaneous evaluation of NotebookLM podcasts and study guides in a US business course, which found higher exam scores in the semester the materials were available (Rezvani, 2026). The two estimates differ greatly in size. Rezvani (2026) reports a difference of roughly three of 15 raw exam points, about 1.5 standard deviations of the comparison semester's scores; our estimate is 2.34 of 50 marks, 0.33 to 0.40 standard deviations. Four features of the designs are likely to account for much of the gap. First, a comparison across semesters with no contemporaneous untreated group attributes every difference between the semesters to the materials. In our data the corresponding comparison, the treated half's mean across cohorts, shows a rise of only 0.73 marks; the difference-in-differences estimate is larger because the untreated half fell, and it nets out shocks common to both halves that a single-group comparison cannot. Second, the assessments differ in stakes. Rezvani's (2026) exams carried 12% of the final grade, whereas our examination carried the whole module mark, so students in both halves of our course prepared with every resource available and new materials competed with intensive revision of all kinds. Third, a 15-point scale converts a few raw points into a large share of the maximum, while ours spreads performance across 50 marks. Fourth, our cohort was high attaining, with module means near 70%, leaving less room above most students than in a course whose comparison semester averaged 62%. Neither estimate should be read as the effect of AI-generated podcasts in general; each is the effect of a particular bundle of materials, compared in a particular way, in a particular assessment regime.

The designs also answer different questions. Rezvani (2026) relates individual engagement to outcomes within the podcast semester, which our data cannot do. Our design compares a treated with an untreated half in the same students and the same year, and examines which part of the attainment distribution moved, which a comparison of means across semesters cannot do.

### *5.3. Verification, credibility and learner trust*

Our second contribution concerns how verification shapes students' engagement. Students' accounts suggest that the label worked as a credibility cue in Metzger's (2007) sense: a low-cost heuristic that substitutes for systematic checking. Unlike most heuristics, it pointed to a person who had done the systematic checking, and students' trust attached to that person rather than to the model. That carries a boundary condition. Students trusted a verifier they knew from tutorials, and a label naming an unknown or anonymous checker may not carry the same weight. Verification works as a signal only when it is legible to the students it addresses.

The same mechanism creates a risk. A label can go on signalling after the checking behind it has become cursory, for instance when the volume of AI material grows faster than the time allocated to checking it, or when a model update changes the character of the output. If students then relax their scrutiny, as the automation literature predicts for aids that appear reliable (Parasuraman & Manzey, 2010), a label that once relieved the judgement burden would hide it from the students it was meant to protect. Because generation is nearly free and verification time grows with volume, this risk rises as institutions scale up AI-generated provision. A record of what was checked, by whom and when, is the means of keeping the label honest.

### *5.4. Evaluating AI learning resources*

Our third contribution concerns evaluation. Claims that AI will widen access to learning concern students who lack the preparation or support that conventional provision assumes. An average treatment effect combines their outcomes with everyone else's and cannot test such claims. Our results show a sharper difficulty: the average effect of the intervention is significant in the full sample and loses significance when three to five students are removed, depending on how standard errors are computed, while the reduction in the share of marks below the upper-second boundary is unchanged in direction and remains significant. An evaluation reporting only the mean would have judged the intervention marginal; a reviewer who probed the mean would have judged it fragile. Neither judgement concerns the students the intervention was designed for. Meta-analyses that pool average effects across studies (Deng et al., 2025) inherit the same limitation.

We therefore propose that evaluations of AI learning resources report, alongside any average effect, three distributional quantities: the change in the share of students below the grade boundaries that matter for progression and employment, the change in dispersion, and the threshold or quantile profile of the effect. Each can be computed from assessment records that institutions already hold.

### *5.5. Implications for practice*

Course teams that issue AI-generated material face a design choice institutional policy rarely name: who checks the output. Our results favour checking by a named tutor before release over leaving checking to students, particularly in large introductory courses, where prior preparation varies most and where verification costs least per student. Verification should be assigned to staff with subject expertise and a teaching relationship with the students, trained to revise prompts and regenerate output rather than to approve first drafts. It should be counted in workload allocations; absorbing it into the unrecorded margins of insecure graduate teaching contracts would convert an efficiency gain into unpaid work. Labels should name the verifier and disclose AI origin, since students' accounts suggest the disclosure itself prompted scrutiny. Course teams should keep a log of verification time and changes per item, so that a decline in checking is visible before it reaches students.

### *5.6. Limitations and future research*

The evidence comes from one course at one institution with one pre-intervention cohort, so parallel trends cannot be tested and cohort differences cannot be excluded. Effort substitution between course halves remains possible; the component correlation moves against it but does not rule it out. The average effect depends on a small number of students, and only the threshold result is robust to removing them. Individual usage logs were not retained, so we cannot relate use to outcomes or test whether the lower-tail gains arose among students who used the materials. The intervention bundled verification with new material in three formats, so the contribution of verification relative to the material itself cannot be identified. The qualitative evidence consists of field notes rather than transcripts and illustrates rather than tests mechanisms. The cohort was high-attaining, with no failures, so whether verified materials reduce failure where failure is common is untested.

The account suggests direct tests. Randomising students or course sections to verified and unverified releases of the same AI material would measure the screening effect on the distribution, which our design cannot isolate. Randomising the label while holding content constant would separate signalling from screening, and comparing named with anonymous verifiers would test the legibility condition. Linking individual usage logs to outcomes, with an encouragement design that shifts use for reasons unrelated to ability, would allow the relation between use and outcomes to be estimated separately by prior attainment, testing whether the lower-tail gains arise among students who used the materials. Measuring students' prior knowledge, language background and study time would allow the lower-tail pattern to be linked to the resources the judgement-burden account names. Replication in courses with higher failure rates would test whether the effect reaches the pass boundary.

## 6. Conclusion

When one half of a course received AI-generated study materials that a named tutor had checked before release, marks in that half held steady while marks in the other half fell, and the protection concentrated among the students who were performing least well. The share of marks below the upper-second boundary fell by 24.7 percentage points relative to the counterfactual; the median and upper tail did not move. Our interpretation is that verification let the materials reach weaker students without imposing on them the judgement burden that AI-generated material otherwise carries. Generative AI tends to widen gaps when learners must judge its output and to narrow them when someone accountable judges it first. Which of the two a course produces depends on where it places that burden, and on whether the institution continues to fund the work of carrying it.

**Declaration of generative AI use**

During the preparation of this work the authors used Claude (Anthropic) to assist with improving the clarity and concision of the writing, checking references against bibliographic databases, and reproducing the statistical analyses as a check on the authors' own code. After using this tool, the authors reviewed, revised and edited all content, verified every reported estimate against the original data, and take full responsibility for the content of the published article. The generative AI tool used to produce the study materials evaluated in this research (Google NotebookLM) is described in Section 3.2.

**Funding**

[Funding details omitted for anonymous review.]

**Declaration of competing interest**

The authors declare that they have no known competing financial interests or personal relationships that could have appeared to influence the work reported in this paper.

**Data availability**

The data are student assessment records and cannot be shared publicly under the terms of the ethics approval.